# Artificial Decision Making Under Uncertainty in Intelligent Buildings

**Magnus Boman**
DSV, Stockholm University and
the Royal Institute of Technology
Electrum 230, SE-164 40 Kista,
Sweden, mab@dsv.su.se

**Paul Davidsson**
Dept. of Software Engineering and
Computer Science
University of Karlskrona/Ronneby
Soft Center, SE-372 25 Ronneby,
Sweden, pdv@ide.hk-r.se

**Håkan L. Younes**
DSV, Stockholm University and
the Royal Institute of Technology
Electrum 230, SE-164 40 Kista,
Sweden, lorens@acm.org

## Abstract

Our hypothesis is that by equipping certain agents in a multi-agent system controlling an intelligent building with automated decision support, two important factors will be increased. The first is energy saving in the building. The second is customer value—how the people in the building experience the effects of the actions of the agents. We give evidence for the truth of this hypothesis through experimental findings related to tools for artificial decision making. A number of assumptions related to agent control, through monitoring and delegation of tasks to other kinds of agents, of rooms at a test site are relaxed. Each assumption controls at least one uncertainty that complicates considerably the procedures for selecting actions part of each such agent. We show that in realistic decision situations, room-controlling agents can make bounded rational decisions even under dynamic real-time constraints. This result can be, and has been, generalized to other domains with even harsher time constraints.

## 1 BACKGROUND

We have taken a multi-agent systems approach to intelligent building control. Our test site Villa Wega in Ronneby, Sweden, is a three-story research laboratory equipped with LonWorks[1] and devices for communicating on the electric grid. Moving from simulation and visualization of events, and of the physical appearance of Villa Wega, to full fielded implementation of hardware control, we must solve a number of difficult problems, some of which we have addressed already (Boman *et al.* 1998). We report here on an attempt at improving results previously obtained, by letting agents use automated decision support when faced with situations in which uncertainty plays a vital role. We will describe our approach using the intelligent building domain throughout the paper, but in the end the usefulness of our approach should be obvious also for other domains.

The objective of the agents is twofold: energy saving, but also increased customer satisfaction through value-added services. As will be shown below, the ability to reason under uncertainty is relevant to both objectives. Energy saving is realized, e.g., by lights being automatically switched off, and room temperature being lowered in empty rooms. Increased customer satisfaction is realized, e.g., by adapting temperature and light intensity according to each person's personal preferences. Our simulations indicate that significant savings, thus far up to 40 per cent, can be achieved (Davidsson & Boman 1998). We now claim that further savings would be possible if agents were to choose autonomously and rationally between action alternatives in real-time situations, rather than resorting to hard-coded action patterns. In our implementation, the use of plan libraries is therefore restricted to static plans. The latter are essentially sequences of primitive operations, which rarely require rearranging.

The multi-agent approach allows for a structure-preserving mapping of the design entities of the application and of the smart equipment of the implementation. It is an open architecture in which agents can be easily configured and re-configured, even dynamically, in the sense of (Cheyer, Martin & Moran 1999). It is also truly distributed, since we make no assumptions about the locations of the agents.

Section 2 very briefly describes some of the agents in Villa Wega. Section 3 motivates the need for agent decision support, and why agents in intelligent buildings must reason under uncertainty. Section 4 summarizes our recent findings in artificial decision making, and the penultimate section gives an example of how they can be used in Villa Wega. We close with conclusions and relatively extensive indications of on-going and future research.

---

[1] See www.echelon.com.

## 2  AN INTELLIGENT BUILDING

In Villa Wega, each electrical device is connected via special purpose hardware nodes to the LonWorks system, allowing the exchange of information over the electrical network. Some of these devices are sensory and some are actuator devices. The sensory device most relevant to this paper is temperature, and to some extent an active badge system. The latter makes it possible to know which persons are in each room at any moment (Harter & Hopper 1994). The actuator devices in the current application are lamps, radiators, and generic mobile devices that can be connected to an arbitrary electrical device, e.g., a coffee machine, or a personal computer.[2] It is possible to switch on and off the device connected to the generic mobile device and to read its state.

These devices interact with, and are controlled by, the multi-agent system (MAS), implemented in April (McCabe & Clark 1995). The sensory devices provide input to the system and the actuator devices occasionally receive instructions from it. This interaction is mediated by a control panel written in Java that translates messages from the MAS to commands understood by the LonWorks system, and *vice versa*. Currently, the entire building environment can be simulated, including the control panel functionality (see Figure 1). In addition, a GUI visualizing the state of the building in terms of temperature, light intensity of the rooms, and the persons present in the rooms has been implemented (see Figure 2).

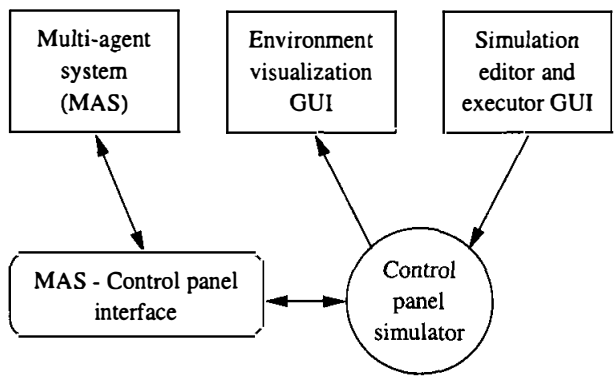

Figure 1: Simulating Villa Wega

There are several categories of agents in the MAS (see, e.g., (Boman *et al.* 1998) for details). We need to consider *Room agents*, which each corresponds to and controls a particular room, with the overall goal of saving energy. Taking into account the preferences of the persons currently (or soon) in the room, it decides what values of the environmental parameters, e.g., temperature and light, are appropriate. *Environmental Parameter (EP) agents* then have access to sensor and actuator devices for reading and changing parameters. For instance, a temperature agent can read the temperature sensor and control the radiators in the room.

## 3  DECISION SITUATIONS AND UNCERTAINTY

Usually, the goal of a Room agent and agents realizing user preferences in the room are conflicting: The Room agent tries to maximize energy savings while other agents try to maximize customer value. In the intelligent building domain, this is the main trade-off. Another type of a conflicting goal situation can be exemplified by the adjustment of temperature in a meeting room in which people with different preferences regarding temperature will meet. The preferences of each person in Villa Wega are encoded in a *Personal Comfort agent*. In good time before a meeting starts, Personal Comfort agents representing each person participating in the meeting negotiate about the temperature. When a particular temperature has been agreed upon, the Room agent RoomMeet delegates a task to an EP agent controlling, e.g., a radiator. Section 5 gives an example where EP agents must choose between various actions affecting radiators and ventilation, in order to achieve their goals. This is a typical decision situation that should be solved by analysis and evaluation of alternatives. This procedure in turn calls for artificial decision making capabilities in the Room agent.

The Villa Wega meeting room is equipped with two 1000W radiators. When the meeting room is empty, the temperature is set to 16°C. There are on average five meetings per week, and the length of each meeting is two hours on average. The persons in Villa Wega have electronic calendars indicating, e.g., which meetings they will participate in. This information is available to RoomMeet, which may use it to plan for the heating of the meeting room in a way that minimizes energy consumption. Here, uncertainty enters the picture: A person might not show up at a meeting. The extent to which a person acts in concordance with her electronic calendar naturally varies, and the probability of her showing up at a particular meeting could be taken into account. Machine learning algorithms notwithstanding, RoomMeet can be informed of precisely who is in the meeting room at a time-point, say, five minutes after the planned start time. This information can be obtained from the smart badge system. RoomMeet can then call for a re-negotiation of the temperature setting with the Personal Comfort agents whose owners are actually at the meeting. After a few seconds, a new setting is obtained. Most likely, the persons at the meeting will not even notice (but possibly appreciate) the slight change of temperature that might occur soon after. RoomMeet does not demonstrate here so much reasoning under uncertainty, as an adaption to a situation deviating from the corresponding expected situation, due to an uncertainty.

---

[2]For information on the so-called ARIGO Switch Station, see www.arigo.de/index-e.htm.

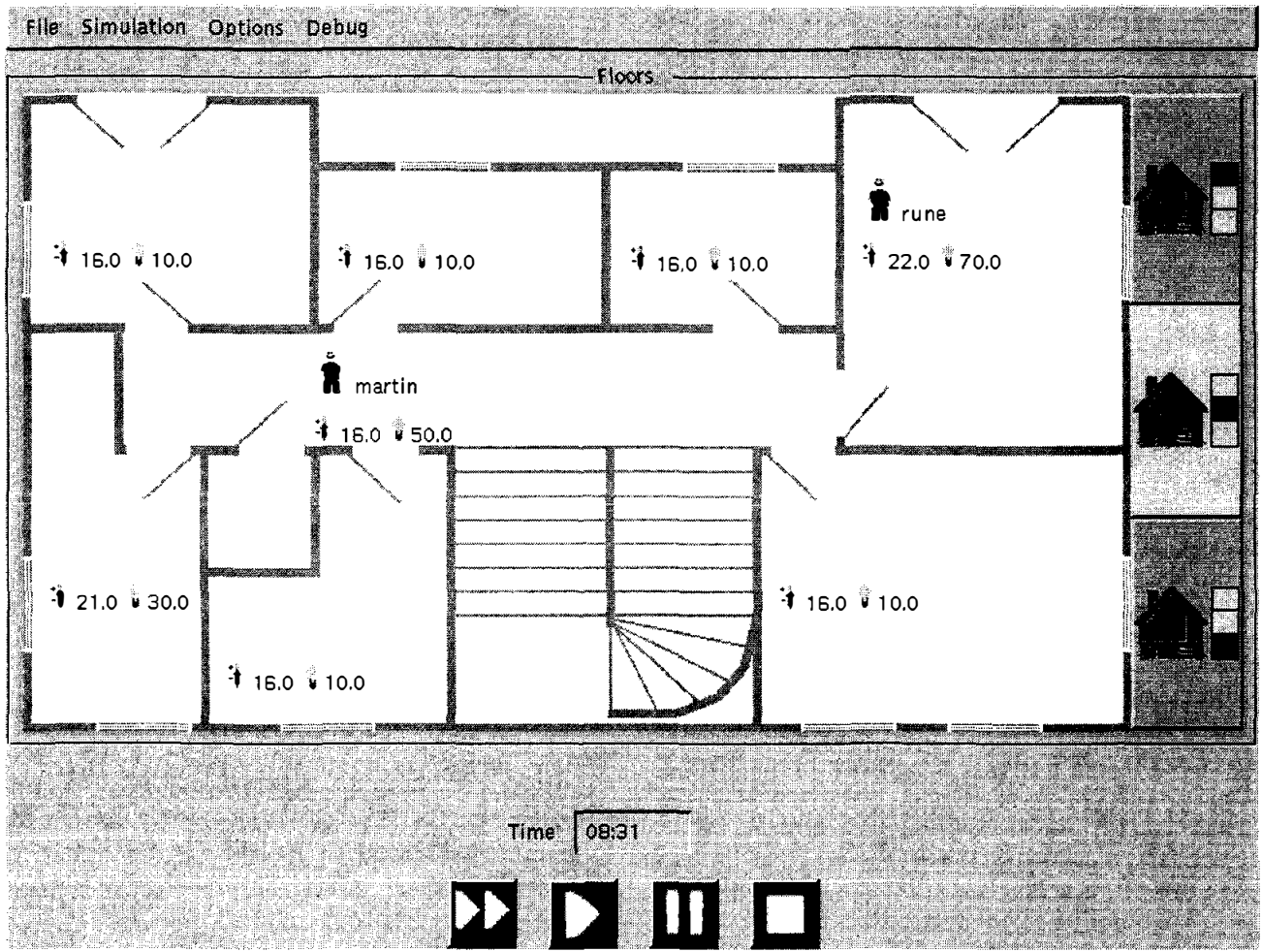

Figure 2: GUI Visualizing the State of the Building

In order to compute the time required to adjust the temperature in a room, the Room agent relies on the thermo-dynamical models described by Incropera and Witt (1990), discretized according to standard procedures described by Ogata (1990). The thermo-dynamical characteristics of a room are described by two constants: the thermal resistance, which captures the heat losses to the environment, and the thermal capacitance, which captures the inertia when heating up/cooling down the entities in the room (see (Davidsson & Boman 1998) for details). We have also made a number of simplifications that affect RoomMeet in our example, such as:

- Outdoor temperature is ignored[3]

- Radiation from the sun is assumed to be negligible

- Radiators have an efficiency of 100 per cent

- Heat produced by persons in a room is ignored

- Heat produced by computers, lamps, and fluorescent tubes is ignored

Relaxing any of the above assumptions means that procedures become more accurate. However, all but the first simplification have a very small effect on any deliberation in RoomMeet. For example, the effects of sun radiation are to a large extent captured by outdoor thermometers, placed on all four sides of the building. Note also that some of the factors would increase the room temperature, while others would decrease it. Outdoor temperature does have a very strong effect, as the difference between a hot summer day and a cold winter day can be more than 50°C in Ronneby. The probabilities considered in our example will therefore be conditioned on the unknown outside temperature, making the example more difficult, but also more interesting.

---
[3]The average 10°C was used in our measurements of energy savings.

## 4 ARTIFICIAL DECISION MAKING

Numerous tools for decision analysis are readily available to human decision makers, aiding them in the structuring and solving of decision situations by means of intuitive GUIs. For an artificial agent, other means to interact with a tool than through a GUI are needed. We have investigated a number of commercial and academic tools for decision analysis, and have concluded that most of the available tools do provide interfaces suitable for agent interaction (Younes 1998). We have run tests on three tools—Netica, SMILE, and DATA Interactive—in the RoboCup domain. Netica and SMILE are based on the algorithm of Shachter and Peot (1992) for finding the optimal policy in an influence diagram. DATA Interactive uses decision trees to represent decision problems, and is based on the *averaging out and folding back* algorithm (Raiffa 1968).

We choose to put the decision analysis functionality of a tool into a *pronouncer* (Boman & Verhagen 1998). This is an authoritative entity external to the agents in the MAS. The agents call upon the pronouncer when they are faced with a decision situation, after which the pronouncer evaluates the given problem and returns an action to the agent. Normally this would be the action that maximizes the expected utility for the agent, but the pronouncer could make use of norms to filter advised actions, in order to account for group utility in a MAS (Boman 1999), or of constraints representing risk aversion (Ekenberg et al. 1999).

The alternative to using a pronouncer would be to have a separate *decision module*, implementing the decision support functionality, in each of the agents. While this would reduce the response time for each decision query, it would also make each agent substantially larger (Younes 1998). The latter can be a problem in our intelligent building application if the agents are to be distributed throughout the building, and not only reside on a central server. In the fielded application, agents may even transport themselves on the electric grid, and hence keeping their size moderate is of in-terest. Moreover, the LonWorks-related hardware installed in Villa Wega also has limitations with respect to memory. The time constrains in our domain are also not severe enough to motivate the use of decision modules.

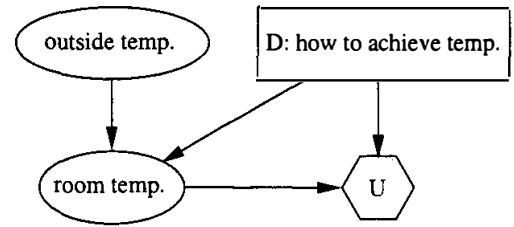

Figure 3: The Decision Problem Represented as an Influence Diagram

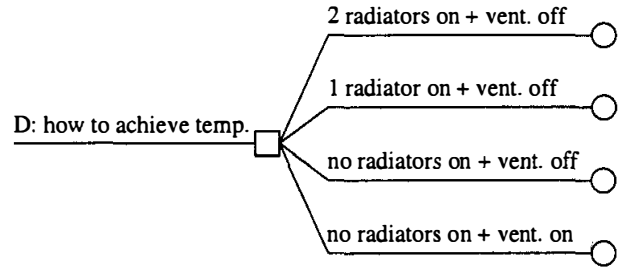

Figure 4: First Level of the Corresponding Decision Tree

## 5 EXAMPLE

The decision problem outlined at the end of Section 3 can be modeled with an influence diagram (Figure 3). The decision node represents four possible actions that the Room agent can perform in order to adjust the room temperature (see Figure 4). The decision will influence the final outcome of the room temperature, but will also influence the utility function since each action has a different effect on the energy consumption. The future outside temperature, which constitutes the uncertainty in this situation, is modeled as a random variable with five possible outcomes ranging from high positive difference between the outside temperature and the desired room temperature to a high negative difference. Finally, the chance node representing the final outcome of the room temperature has three possible outcomes: The temperature is higher than desired, the temperature is desirable, or the temperature is lower than desired.

We have run 10000 set/evaluate-runs on the given influence diagram using Netica and SMILE in order to determine if it is feasible to use either of them for decision support in the Villa Wega agent system. We have also transformed the influence diagram into a de-cision tree, which we then have evaluated using a basic decision tree evaluator (BDTE) implemented by ourselves, based on the averaging out and folding back algorithm. The performance measure used in all tests was the time it takes to first set all values in the model, and then evaluate the whole model. The platform was a 167 MHz Sparc Ultra Creator running Solaris.

In our implementations we do not allow the agents to formulate decision situations on the fly, since it would be extremely difficult for them to do so (Boman 1999). Instead, the pronouncers contain template models designed in advance for decision situations that can be presumed to occur. Each template determines the structure of a certain problem, while it is up to the agents to specify the values. How these values are set can vary with the implementation. In addition to the usual way of adding value nodes to an influence diagram, various models of prediction can be adopted, see, e.g., (Ygge & Akkermans 1997).

As Table 1 shows, even the slowest tool—SMILE—solves the decision problem at hand in less than ten milliseconds. With this short response time, the agents could make extensive use of the decision support provided by a pronouncer without any noticeable degradation of the system. This is an important point, as many researchers have taken a stance against real-time decision support for artificial agents. We have therefore made an effort to not only demonstrate that the use of pronouncers is feasible in many domains, but also to implement them.

Table 1: Means and Standard Deviations (in milliseconds) for 10000 Runs

| TOOL | MEAN | STD.DEV. |
|---|---|---|
| BDTE | 0.03 | 0.01 |
| Netica | 2.12 | 0.17 |
| SMILE | 7.31 | 1.48 |

## 6 CONCLUSIONS AND FUTURE WORK

By investigating a non-trivial prototypical example, we have demonstrated how agents part of a multi-agent system controlling parts of an intelligent building can reason under uncertainty. The agents make calls to a pronouncer which provides extremely fast decision support by evaluating the input (a decision tree or an influence diagram) and returning the best action. We have not limited the scope of this support by assuming a particular decision rule, e.g., the principle of maximizing the expected utility. On the contrary, we are interested in how different classical extensions of the principle, such as risk attitudes (Ekenberg et al. 1999), group rationality (Boman 1999), and meta-rules (Laskey & Lehner 1994) affect agent behavior.

What is meant by the best action is determined in part by the nature of the social space (sometimes called an artificial ecosystem) that the agent is in. For instance, one agent can be part of several coalitions, each of which constrains its actions considerably, while another agent is more individualistic. We are currently investigating the use of norms to achieve socially intelligent behavior in a number of domains, including intelligent buildings. Analogous to our tolerant view on decision rules, our view on technical norms is that various implementations (e.g., active/passive norms) are worthy of study, and that meta-norms (Axelrod 1986) must come into play.

The high speed of the implemented pronouncers make several extensions of their functionality possible. The commercial tools investigated have attractive features that cannot be used in domains with severe time constraints, e.g. RoboCup, but which may prove most useful in intelligent buildings. One such extension is to vague and imprecise data. The precise values handled by the pronouncers described in this paper are awkward and unrealistic in many situations. We are therefore investigating the possible employment of our algorithms for evaluation of situations with imprecise values, originally developed for management systems and human decision support (see, e.g., (Ekenberg, Danielson & Boman 1996), (Ekenberg, Danielson & Boman 1997)), in artificial decision making.

It must always be possible to over-rule the decisions of the agents in the MAS by physical interaction with the electrical equipment. For instance, even if an EP agent has decided that the light in a room should be on, it must be possible for a person to turn off the light using the switch in the actual room. These constraints are, of course, not hard-wired into the MAS and can be changed easily. We have studied the problem of manual overrides and its effects on agent autonomy (Verhagen & Boman 1999), and intend to pursue this research in the domain of intelligent buildings.

As the above listing of ongoing and future related research indicates, intelligent buildings is not the only domain of interest to the proposed methods, algorithms, and implementations. We hope to have made clear that by choosing a prototypical example our intention was only to make a presentation abstract enough to allow for straightforward mappings to other domains.

## References

Axelrod, R. 1986. An evolutionary approach to norms. *American Political Science Review* 80(4):1095–1111.

Boman, M., and Verhagen, H. 1998. Social intelligence as norm adaptation. In Dautenhahn, K., and Edmonds, B., eds, *SAB'98 Workshop on Socially Situated Intelligence*, 17–25. Technical Report, Centre for Policy Modelling.

Boman, M.; Davidsson, P.; Skarmeas, N.; Clark, K.; and Gustavsson, R. 1998. Energy saving and added customer value in intelligent buildings. In Nwana, H. S., ed., *Proc PAAM'98*, 505–516.

Boman, M. 1999. Norms in artificial decision making. *Artificial Intelligence and Law Journal* 7(1).

Cheyer, A.; Martin D.; and Moran D. 1999. The open agent architecture: A framework for building distributed software systems. *Applied Artificial Intelligence* 13(1–2).

Davidsson, P., and Boman, M. 1998. Energy saving and value added services: Controlling intelligent buildings using a multi-agent systems approach. In *DA/DSM Europe DistribuTECH*, PennWell.

Ekenberg, L.; Danielson, M.; and Boman, M. 1996. From local assessments to global rationality. *Intl Journal of Intelligent Cooperative Information Systems* 5(2&3):315–331.

Ekenberg, L.; Danielson, M.; and Boman, M. 1997. Imposing security constraints on agent-based decision support. *Decision Support Systems Intl Journal* 20:3–15.

Ekenberg, L.; Boman, M.; and Linnerooth-Bayer, J. 1999. General risk constraints. *Journal of Risk Research*. Under revision.

Harter, A., and Hopper A. 1994. A distributed location system for the active office. *IEEE Network* 8(1).

Incropera, F. P., and Witt, D. P. 1990. *Fundamentals of Heat and Mass Transfer*. Wiley and Sons, third edition.

Laskey, K. B., and Lehner, P. E. 1994. Metareasoning and the problem of small worlds. *IEEE Transactions on Systems, Man, and Cybernetics* 24(11).

McCabe, F. G., and Clark, K. L. 1995. April: Agent process interaction language. In Wooldridge M. J. and Jennings N. R., eds., *Intelligent Agents*. Springer-Verlag. 324–340. LNAI 890.

Ogata, K. 1990. *Modern Control Engineering*. Prentice-Hall, second edition.

Raiffa, H. 1968. *Decision Analysis*. Reading, Mass.: Addison-Wesley.

Shachter, D. R., and Peot, M. A. 1992. Decision making using probabilistic inference methods. In Dubois D., Wellman M. P., D'Ambrosio B., and Smets P., eds., *Proc UAI'92*. Morgan Kaufmann.

Verhagen, H., and Boman, M. 1999. Adjustable autonomy, norms and pronouncers. In *Proc AAAI Spring symposium on adjustable autonomy*. Stanford Univ technical report. Forthcoming.

Ygge, F., and Akkermans H. 1997. Making a case for multi-agent systems. In Boman M. and Van de Velde W, eds., *Multi-Agent Rationality*. Springer-Verlag. 156–176. LNAI 1237.

Younes, H. L. 1998. Current tools for assisting intelligent agents in real-time decision making. Master's thesis, Royal Institute of Technology and Stockholm University, Stockholm, Sweden. No. 98-x-073.